\pdfoutput=1

\documentclass[11pt,a4paper]{article}
\usepackage[hyperref]{acl2023}
\usepackage{times}
\usepackage{latexsym}
\usepackage{bm}
\usepackage{amsmath}
\usepackage{url}
\usepackage{makecell}
\usepackage{multirow}
\usepackage{subfigure}
\usepackage{graphicx} 
\usepackage{ulem}
\usepackage{colortbl}
\usepackage{arydshln}
\usepackage[ruled,linesnumbered]{algorithm2e}

\definecolor{mypink}{rgb}{.99,.91,.95}
\aclfinalcopy
\def\aclpaperid{5}
\usepackage{booktabs}
\usepackage{multirow,multicol}

\title{Bag of Tricks for Effective Language Model Pretraining and Downstream Adaptation: A Case Study on GLUE}

\author{Qihuang Zhong$^{\diamondsuit, \Re \includegraphics[scale=0.15]{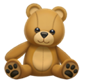}}$,
\ Liang Ding$^{\Re  \includegraphics[scale=0.15]{first.png}}$,
\ Keqin Peng$^{\natural}$
\\ 
\ \textbf{
\ Juhua Liu$^{\diamondsuit}$, 
\ Bo Du$^{\diamondsuit}$,  
\ Li Shen$^{\Re}$,
\ Yibing Zhan$^{\Re}$, 
\ Dacheng Tao$^{\Re}$} \\
\ $^{\diamondsuit}$Wuhan University
\ $^{\Re}$JD Explore Academy, JD.com Inc. 
\ $^{\natural}$Beihang University\\
\includegraphics[scale=0.15]{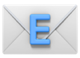} \texttt{zhongqihuang@whu.edu.cn}, \texttt{dingliang1@jd.com}
}
\begin{document}
\maketitle
\renewcommand{\thefootnote}{\fnsymbol{footnote}}
\footnotetext{\includegraphics[scale=0.15]{first.png} Equal contribution. Work was done when Qihuang and Keqin were interning at JD Explore Academy.}
\begin{abstract}
This technical report briefly describes our JDExplore d-team’s submission \textbf{Vega v1} on the General Language Understanding Evaluation (GLUE) leaderboard\footnote{\url{https://gluebenchmark.com/leaderboard/}}, where GLUE is a collection of nine natural language understanding tasks, including question answering, linguistic acceptability, sentiment analysis, text similarity, paraphrase detection, and natural language inference.
\texttt{\bf [Method]}
We investigate several effective strategies and choose their best combination setting as the training recipes. As for model structure, we employ the vanilla Transformer with disentangled attention as the basic block encoder. For self-supervised training, we employ the representative denoising objective (i.e., replaced token detection) in phase 1 and combine the contrastive objective (i.e., sentence embedding contrastive learning) with it in phase 2. During fine-tuning, several advanced techniques such as transductive fine-tuning, self-calibrated fine-tuning, and adversarial fine-tuning are adopted. 
\texttt{\bf [Results]}
According to our submission record (Jan. 2022), with our optimized pretraining and fine-tuning strategies, our 1.3 billion model sets new state-of-the-art on 4/9 tasks, achieving the best average score of 91.3. Encouragingly, our Vega v1 is the first to exceed powerful human performance on the two challenging tasks, i.e., SST-2 and WNLI.
We believe our empirically successful recipe with a bag of tricks could shed new light on developing efficient discriminative large language models.
% The model will be released through the OmniForce Platform\footnote{OmniForce Platform will be launched by JD Explore Academy}.
\end{abstract}

\begin{figure}[htb]
    \centering
    \includegraphics[width=0.47\textwidth]{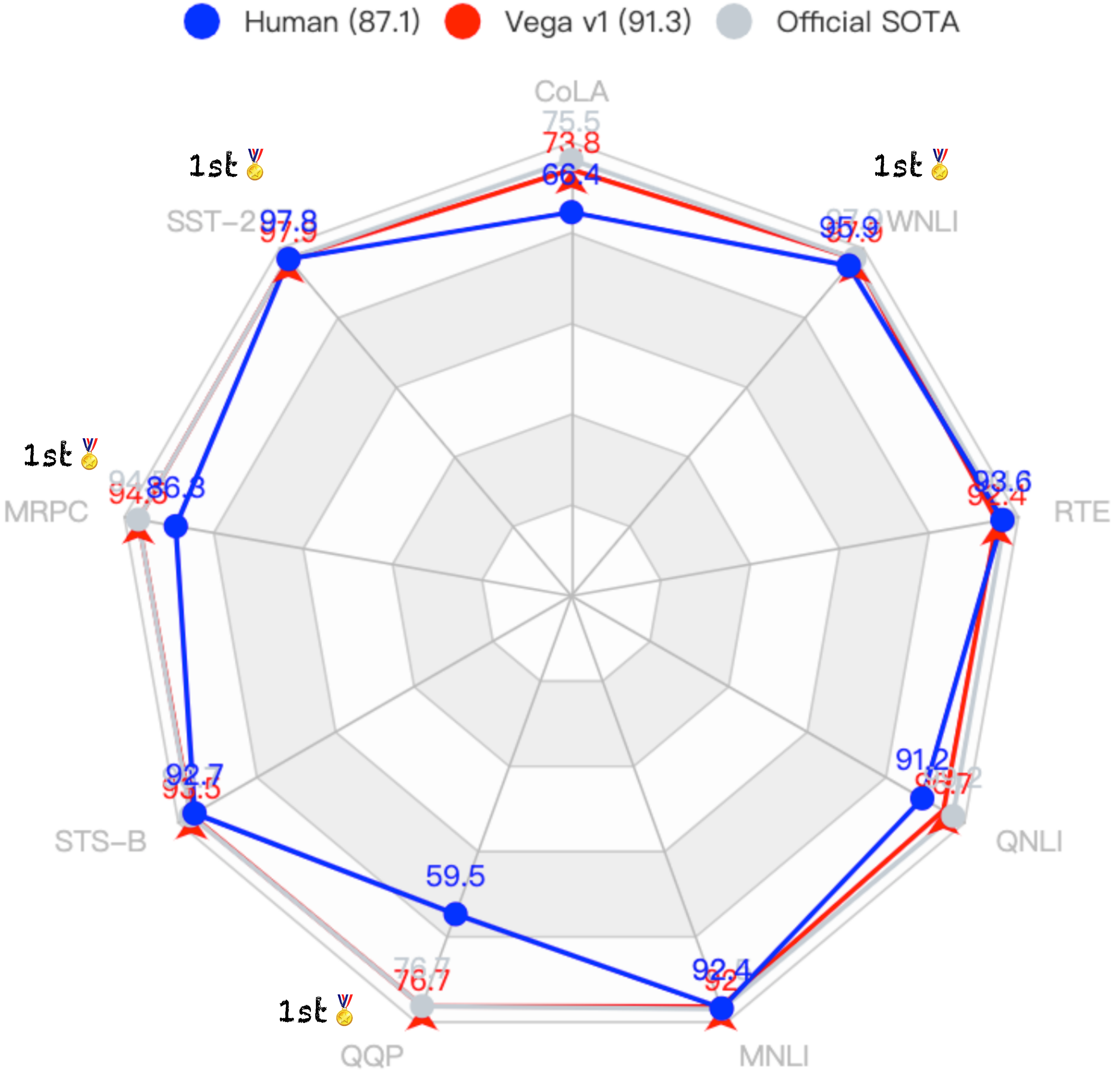}
    \caption{Vega v1 achieves 4 state-of-the-art records out of 9 tasks among all submissions, winning the best average score of 91.3, and significantly outperforming the GLUE Human performance.}
    \label{fig:radia}
\end{figure}

\section{Introduction}
Pretrained language models (PLMs)~\cite{devlin-etal-2019-bert,liu2019roberta,deberta,joshi2020spanbert,sun2019ernie,GPT3,t5} are widely used in the community of natural language processing and have achieved remarkable success in numerous downstream tasks of natural language understanding (NLU), such as sentiment analysis~\cite{liu2021unified,wang2022contrastive,zhong2022knowledge}, intent detection~\cite{kim2016intent,wu2020slotrefine} and reasoning~\cite{qu2022interpretable}. These PLMs share a common principle of performing self-supervised learning with massive easy-to-acquire unlabelled text corpora during the pretraining stage and effectively fine-tuning on downstream tasks. In such a context, the general language understanding evaluation (GLUE,~\citealt{wang2018glue}) benchmark has emerged as the leading evaluation standard for the pretrained language model community, where most high-performing models (e.g., T5~\cite{t5})  on its leaderboard provide valuable insights and best practices for future research and applications.

We recently submitted our 1.3B \textbf{Vega} v1 model to the GLUE leaderboard and, as seen in Figure~\ref{fig:radia}, obtained state-of-the-art records on 4 out of 9 tasks, sitting atop the leaderboard as of January 1, 2022, with an average score of 91.3. 
More encouragingly, our Vega v1 is the first to exceed powerful human performance on the two challenging tasks, i.e., SST-2~\cite{socher2013recursive} and WNLI~\cite{levesque2012winograd}. This technical report briefly describes how we build our powerful model under a certain parameter budget, i.e., 1.3B, from different aspects, including backbone framework (\S\ref{subsec:backbone}), efficient pretraining processes (\S\ref{subsec:pretrain}), and effective downstream adaptation approaches (\S\ref{subsec:downstream}).
To achieve efficient and sufficient pretraining, we replace the widely-used masked language modeling (MLM,~\citealt{devlin-etal-2019-bert}) with two simple but effective objectives, i.e., denoising and contrastive objectives. The denoising objective~\cite{yamaguchi2021frustratingly} aims to improve the data efficiency and save training costs, while the contrastive objective involves leveraging contrastive learning~\cite{gao2021simcse} to learn better sentence representations. 
For downstream adaptation, we focus on two common problems, i.e., domain discrepancy and over-fitting, and adopt several effective fine-tuning methods, such as self-calibrated transductive fine-tuning and adversarial fine-tuning, to achieve better performance and generalization.

The rest of this paper is organized as follows. In Section~\ref{sec:app}, we introduce the major utilized approaches. Then, Section~\ref{sec:exp} reports and discusses our evaluation results. Finally, we conclude our study in Section~\ref{sec:con}.

\section{Approaches}
\label{sec:app}
In this section, we describe the main techniques in our Vega v1 model, including the backbone framework in~\S\ref{subsec:backbone}, the efficient pretraining approaches in~\S\ref{subsec:pretrain}, and the effective downstream adaptation technique in~\S\ref{subsec:downstream}.

\subsection{Backbone Framework}
\label{subsec:backbone}
In recent years, the Transformer~\cite{transformer} has become the \textit{de-facto} standard for neural language modeling, and has achieved great success in the field of large-scale language model pretraining~\cite{devlin-etal-2019-bert,t5,GPT3,liu2021complementarity,zhong2022e2s2,wang2022understanding,Zan2022PTvsRI,zan-etal-2022-vega}. In terms of network architectures, the Transformer-based PLMs can be classified into three groups: decoder-only models (e.g., GPT-3~\cite{GPT3}), encoder-only model (e.g., BERT~\cite{devlin-etal-2019-bert}) and encoder-decoder models (e.g., T5~\cite{t5}). As the encoder-only models have an overwhelming advantage over the existing methods on the GLUE leaderboard, we train our large model in an encoder-only manner to facilitate downstream language understanding tasks. In particular, the self-attention mechanism adopted in Transformer can effectively encode the content information, but unfortunately lacks a natural way to encode word position information~\cite{dufter2022position,ding2020self}. Thus, following~\citet{deberta}, we replace the vanilla self-attention mechanism in Transformer with a disentangled attention strategy. According to our Vega v1 parameter budget -- 1.3 Billion, we empirically set the model as follows: 48 layers, 1,536 as the hidden layer size, an FFN of size 6,144, 24 heads, and 64 as the head size.

\subsection{Efficient Pretraining}
\label{subsec:pretrain}
\begin{figure*}[t]
    \centering
    \includegraphics[width=0.98\textwidth]{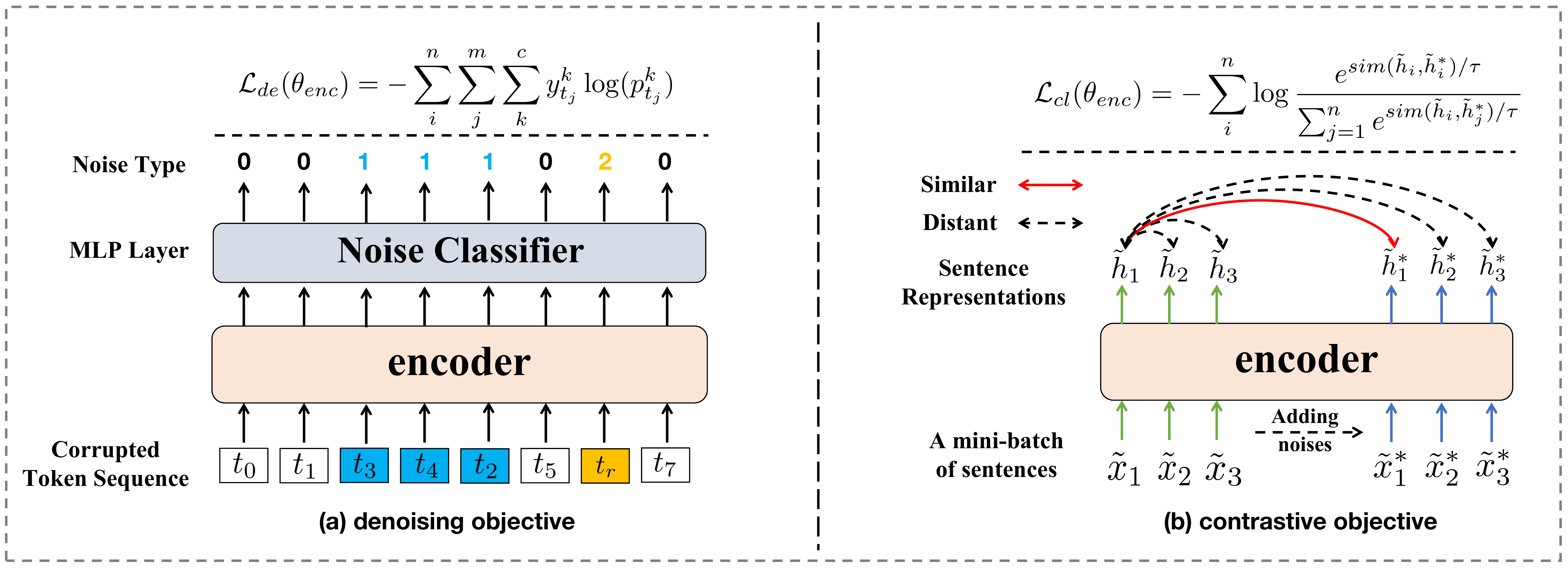}
    \caption{\textbf{Illustrations of denoising (a) and contrastive (b) objectives}. In \textbf{(a)}, the token sequence is obtained by corrupting the original input tokens $\{t_0,t_1,t_2,t_3,t_4,t_5,t_6,t_7\}$, where the boxes in \textcolor[rgb]{0,0.69,0.94}{blue} denote the shuffled tokens and the \textcolor[rgb]{1,0.75,0}{yellow} box denotes the randomly replaced token. $n$, $m$, and $c$ are the number of sentences in a mini-batch, the length of the token sequence, and the classes of noise type (1 for shuffle, 2 for random replacement, and 0 for others) respectively. $y$ and $p$ are the ground truths and predictions of the noise type. In \textbf{(b)}, $\tilde{x}$ and $\tilde{x}^*$ denote the original sentence and corrupted sentence respectively. $\tilde{h}$ and $\tilde{h}^*$ are the corresponding sentence representations.}
    \label{fig:objective}
\end{figure*}

One key component of language model pretraining is the pretraining objective. Most of the prior existing pretrained models are usually based on the masked language modeling (MLM) objective~\cite{devlin-etal-2019-bert,liu2019roberta} or its variants~\cite{sun2019ernie,joshi2020spanbert}. However, the MLM is usually criticized as being inefficient, as it incurs a substantial compute cost (top-layer vocabulary-dimension embedding) but only produces supervision signals at a small proportion of positions (usually 15\%)~\cite{clark2020electra}. To this end, we propose two more efficient pretraining objectives (i.e., denoising and contrastive objectives) for effectively training our Vega v1, which are illustrated in Figure~\ref{fig:objective}. Specifically, the denoising objective and contrastive objective are expected to capture the local token-level and global sentence-level knowledge, respectively.
In this part, we first compare our objectives with the vanilla MLM, and then detailed introduce the two-stage pretraining algorithm used in our Vega v1.

\subsubsection{Pretraining Objectives}
\paragraph{Masked Language Modeling} MLM randomly selects a subset (usually 15\%) of tokens from a sentence and replaces them with a special mask token, i.e., \texttt{[MASK]}. Then, the MLM trains a 
model to predict a particular token (from the vocabulary space) that has been replaced with a \texttt{[MASK]} placeholder given its surrounding context. 

\paragraph{Denoising Objective} Instead of using the MLM, motivated by the success of ELECTRA~\cite{clark2020electra} and other similar works~\cite{yamaguchi2021frustratingly,alajrami2022does}, we use a more efficient replaced token detection (RTD) as an alternative. The principle of RTD is to manually corrupt the sentence and encourage the model to denoise the corrupted sentence. In practice, for each sentence, we replace 10\% of tokens with shufﬂed ones from the 
same sequence and another 5\% of tokens with 
random ones from the vocabulary. Then, the model is forced to learn the linguistic knowledge for detecting the shuffled or replaced tokens among the supervisions of all input tokens. It is noteworthy that, different from the MLM that predicts the token from the vocabulary space, our denoising objective is a \textit{ternary classification} task, aiming at identifying whether a token in the input sequence has been shufﬂed or randomly replaced or not, which is more sample-efficient and saves large training costs.

\paragraph{Contrastive Objective} In addition to the above denoising objective that focuses on token-level linguistic information learning, we further introduce a contrastive objective to improve the sentence-level representation learning ability of our Vega v1. Specifically, the motivation of this objective is that many prior studies~\cite{li2020sentence,gao2021simcse} have found that Transformer-based models always induce a non-smooth anisotropic semantic space of sentences, which harms the performance of sentence representation. 

To alleviate this problem, our contrastive objective adopts the contrastive learning technique on sentence representation. The two key problems of contrastive learning are 1) how to construct positive instances; and 2) how to obtain sentence representations. For problem 1), we simply use the original sentence and its corresponding noising sentence (corrupted with the RTD in the denoising objective) as the positive instance pair. As for problem 2), we can directly use the \texttt{[CLS]} embedding, or simply use basic pooling operations (e.g., mean pooling) to process the hidden representations on the last layer, as the sentence representation. Notably, in the preliminary experiments, we found that there is a slight difference between both methods, whereas the former one is lastly adopted for simplicity.

\subsubsection{Two-stage Pretraining Strategy}
Despite the remarkable performance improvement, the contrastive objective leads to external computation overhead, as it requires two forward-propagation processes. Thus, to better trade off the computation costs and performance, we perform the pretraining process in a two-stage manner. Specifically, in phase 1, we only employ the denoising pretraining objective to ensure the model quickly learns the basic token-level linguistic knowledge from pretraining data. In phase 2, we combine both objectives and continue pretraining the model to further encourage it to fully exploit the knowledge and learn better sentence representation. Note that such fine-to-coarse multi-stage training recipe has shown effective performance in many tasks, e.g., classification~\cite{mcdonald2007structured} and translation~\cite{ding-etal-2021-progressive}.

\subsection{Effective Downstream Adaptation}
\label{subsec:downstream}
In addition to the above efficient and sufficient pretraining methods, we also design some useful fine-tuning strategies for effectively adapting our Vega v1 to downstream tasks. 
Here, we focus on two main problems that hinder the adaptation performance of our Vega v1. Specifically, \textit{\textbf{(i) Domain gap}}. The first concerns the domain gaps between the training and test sets, which lead to poor performance on target test sets. \textit{\textbf{(ii) Over-fitting}}. Due to the limited downstream training data or its hard-to-learn ability, the fine-tuning model usually suffers from the over-fitting problem and shows poor model generalization.

Note that in addition to the strategies listed below, we have also designed and implemented other methods from different perspectives to improve the generalization and efficiency of models, e.g. the FSAM optimizer for PLMs~\cite{Zhong2022ImprovingSM}, PromptTuning with reusing existing prompts~\cite{zhong2022panda}, SparseAdapter~\cite{He2022SparseAdapterAE}, and continued training with downstream data~\cite{Zan2022BridgingCG}. Although these approaches can help to some extent, they do not provide complementary benefits compared to the listed approaches, so they are not described here.

\paragraph{Transductive Fine-tuning}
\label{subsubsec:tf}
\begin{algorithm}[t]
\KwIn{Finetuned (FT) Model $M_0$, ~~~~~~~~~~~~~~~~Training set $D_s$, Test set $D_t$}
\KwOut{Transductively FT Model $M_o$}
        $i:=0$\\
        \While {not convergence}{
      Estimate $D_t$ with $M_i$ and get $D^{M_i}_t$\\
      Tune $M_i$ on $D_s \cup D^{M_i}_t$ and get $M_{i+1}$\\ 
        $i := i + 1$\\      
        }
        $M_o:=M_t$
\caption{Transductive Finetuning} 
\label{alg:1}
\end{algorithm}
Regarding the domain or linguistic style gap between the training and test sets (the problem \textit{\textbf{(i)}}), we adopt a transductive fine-tuning strategy to improve the target domain performance, which is a common practice in machine translation evaluations~\cite{wu-etal-2020-tencent,ding2021usyd} and some domain adaptation applications~\cite{liu2020semitext}. Specifically, let the training set (denoted as $D_s$) be the source domain and the test set (denoted as $D_t$) be the target domain, the key idea of transductive fine-tuning is to transform the target domain into the source domain space with the well-performed model $M_0$ (trained on the source domain), and obtain the generated synthetic dataset. Then, the model is further tuned on this synthetic dataset.
The proposed transductive fine-tuning technique is shown in Algorithm~\ref{alg:1}. Whether we should conduct transductive fine-tuning depends on the practical downstream performance achieved.

\begin{algorithm}[t]
\KwIn{Finetuned (FT) Model $M_0$, ~~~~~~~~~~~~~~~~Training set $D_s$, Test set $D_t$, External data $D^*$}
\KwOut{Transductively FT Model $M_o$}
        Train a language model on $D_s$\\
        Select the data $D^*_s$ (similar to $D_s$) from $D^*$ via the trained language model\\
        Re-label $D^*_s$ with $M_0$ and get $\Tilde{D}^*_s$\\
        \uIf{$D^*_s$ is labeled}{
        Select the samples from $\Tilde{D}^*_s \cap D^*_s$ as the calibrated data
        }\Else{
        Use $\Tilde{D}^*_s$ as the calibrated data
        }
        % Calibrate $D^*_s$ with $M_0$\\
        Tune $M_0$ on the calibrated data and get $M^{'}_0$\\
        $i:=0$\\
        \While {not convergence}{
      Estimate $D_t$ with $M^{'}_i$ and get $D^{M^{'}_i}_t$\\
      Tune $M^{'}_i$ on $D_s \cup D^{M^{'}_i}_t$ and get $M^{'}_{i+1}$\\ 
        $i := i + 1$\\      
        }
      $M_o:=M^{'}_t$
\caption{Self-calibrated Finetuning} 
\label{alg:2}
\end{algorithm}
\paragraph{Self-calibrated Fine-tuning}
\label{subsubsec:scf}
The aforementioned transductive fine-tuning works well when the well-performed finetuned model $M_0$ is available, but performs poorly when it is hard to train the base model $M$. For example, there is usually limited labeled data in the downstream training set (low-resource settings), which hinders the effective training of $M_0$. A natural way is to leverage the external (easy-to-obtain but low-quality) labeled data or even unlabeled data (denoted as $D^*$) from a similar source domain for training the $M_0$. Obviously, it is sub-optimal or even impractical to directly apply the external noisy data to obtain the satisfactory $M_0$.
\begin{table*}[ht]
\caption{\textbf{Results obtained on the GLUE test sets}, which are scored by the GLUE evaluation server. We obtained the results from \url{https://gluebenchmark.com/leaderboard} on January 1, 2022. The best results (except those of human baselines) are shown in bold.}
\scalebox{0.73}{
\centering
\begin{tabular}{lcccccccccccccc}
\toprule
\multicolumn{1}{c}{} & CoLA & SST-2 & \multicolumn{2}{c}{MRPC} & \multicolumn{2}{c}{STS-B} & \multicolumn{2}{c}{QQP} & \multicolumn{2}{c}{MNLI} & QNLI & RTE & WNLI &  \\ \cmidrule{2-14}
\multirow{-2}{*}{\textbf{Models}} & \textit{Mcc.} & \textit{Acc.} & \textit{F1} & \textit{Acc.} & \textit{Pcor.} & \textit{Scor.} & \textit{F1} & \textit{Acc.} & \textit{m.} & \textit{mm.} & \textit{Acc.} & \textit{Acc.} & \textit{Acc.} & \multirow{-2}{*}{Score} \\ \midrule \midrule
\textbf{GLUE Human Baselines} & 66.4 & 97.8 & 86.3 & 80.8 & 92.7 & 92.6 & 59.5 & 80.4 & 92.0 & 92.8 & 91.2 & 93.6 & 95.9 & 87.1 \\ \hdashline
\textbf{T5} & 71.6 & 97.5 & 92.8 & 90.4 & 93.1 & 92.8 & 75.1 & 90.6 & 92.2 & 91.9 & 96.9 & 92.8 & 94.5 & 90.3 \\
\textbf{ALBERT + DAAF + NAS} & 73.5 & 97.2 & 94.0 & 92.0 & 93.0 & 92.4 & 76.1 & 91.0 & 91.6 & 91.3 & 97.5 & 91.7 & 94.5 & 90.6 \\
\textbf{MacALBERT + DKM} & 74.8 & 97.0 & \textbf{94.5} & \textbf{92.6} & 92.8 & 92.6 & 74.7 & 90.6 & 91.3 & 91.1 & 97.8 & 92.0 & 94.5 & 90.7 \\
\textbf{DeBERTa / TuringNLRv4} & 71.5 & 97.5 & 94.0 & 92.0 & 92.9 & 92.6 & 76.2 & 90.8 & 91.9 & 91.6 & \textbf{99.2} & 93.2 & 94.5 & 90.8 \\
\textbf{DeBERT + CLEVER} & 74.2 & 97.5 & 93.3 & 91.0 & 93.4 & 93.1 & 76.4 & 90.9 & 92.1 & 91.7 & 96.7 & 93.1 & 96.6 & 91.0 \\
\textbf{StructBERT + CLEVER} & 75.3 & 97.7 & 93.9 & 91.9 & 93.5 & 93.1 & 75.6 & 90.8 & 91.7 & 91.5 & 97.4 & 92.5 & 95.2 & 91.0 \\
\textbf{ERNIE} & \textbf{75.5} & 97.8 & 93.9 & 91.8 & 93.0 & 92.6 & 75.2 & 90.9 & 92.3 & 91.7 & 97.3 & 92.6 & 95.9 & 91.1 \\
\textbf{TuringNLR v5} & 72.6 & 97.6 & 93.8 & 91.7 & \textbf{93.7} & \textbf{93.3} & 76.4 & 91.1 & \textbf{92.6} & \textbf{92.4} & 97.9 & \textbf{94.1} & 95.9 & 91.2 \\ \midrule
\textbf{Vega v1 (Ours)} & 73.8 & \textbf{97.9} & \textbf{94.5} & \textbf{92.6} & 93.5 & 93.1 & \textbf{76.7} & \textbf{91.1} & 92.1 & 91.9 & 96.7 & 92.4 & \textbf{97.9} & \textbf{91.3} \\
\bottomrule
\end{tabular}
}
\label{table_main}
\end{table*}

To this end, we further present a new self-calibrated strategy to boost the effectiveness of transductive fine-tuning. In practice, the self-calibrated strategy contains a three-stage process: 
\begin{itemize}
    \item 1) Selecting the high-quality sub-dataset $D^*_s$ (similar to $D_s$) from $D^*$ estimated by a language model (e.g., KenLM~\cite{heafield2011kenlm}) trained on $D_s$;
    \item 2) Calibrating $D^*_s$ via adopting the initialized $M_0$ to re-label the data;
    \item 3) Tuning the $M_0$ on the calibrated external data to obtain the well-performed $M^{'}_0$ and further performing the transductive fine-tuning.
\end{itemize}
% 1) selecting the data $D^*_s$ (similar to $D_s$) from $D^*$ with a language model (e.g., KenLM~\cite{heafield2011kenlm}) trained on $D_s$; 2) calibrating $D^*_s$ via adopting the initialized $M_0$ to re-label the data; 3) tuning the $M_0$ on the calibrated external data to obtain the well-performed $M^{'}_0$ and further performing the transductive fine-tuning.
More specifically, for process 2), we first enforce $M_0$ to make the prediction for each sample in $D^*_s$, which is used as the sample's pseudo label, and further construct a new corpus $\Tilde{D}^*_s$. Then, we select the samples from $\Tilde{D}^*_s \cap D^*_s$, i.e., the samples whose pseduo labels are consistent with the prior labels in $D^*_s$, as the calibrated data. Notably, if $D^*$ is a unlabeled corpus, we directly use the $\Tilde{D}^*_s$ as the calibrated data
The proposed self-calibrated fine-tuning technique is shown in Algorithm~\ref{alg:2}.
% If $D^*$ is a (usually noisy) labeled corpus, we first enforce $M_0$ to make the prediction for each sample in $D^*_s$, and select the samples whose predictions are consistent with the previous labels as the calibrated data. Otherwise, we directly employ the predictions of $M_0$ on $D^*_s$ as its pseudo labels and refer to this relabeled $D^*_s$ as the calibrated data. 

\paragraph{Adversarial Fine-Tuning}
Regarding the problem \textit{\textbf{(ii)}}, we are inspired by many prior adversarial training methods~\cite{Miyato2019VirtualAT,Jiang2020SMARTRA,zhang2022improving} and adopt an adversarial fine-tuning to alleviate the over-fitting problem. In practice, to further improve the training stability, we follow \newcite{deberta} and apply the perturbations to the normalized word embeddings when tuning our Vega foundation model on downstream tasks, where we first normalize the embedding vectors into stochastic vectors and then apply the perturbations to the normalized embedding vectors. Similar to the observations of~\newcite{deberta}, we also empirically find that such a normalization improves the generalization and performance of the ﬁne-tuned models on several downstream tasks.

\section{Experiments}
\label{sec:exp}
\subsection{Implementation}
For pretraining, we follow many prior works~\cite{liu2019roberta,deberta} and use Wikipedia\footnote{\url{https://dumps.wikimedia.org/enwiki/}} (the English Wikipedia dump, 10 GB), BookCorpus~\cite{zhu2015aligning}\footnote{\url{https://github.com/butsugiri/homemade_bookcorpus}} (6 GB), OpenwebText\footnote{\url{http://Skylion007.github.io}} (38 GB), Stories\footnote{\url{https://github.com/tensorﬂow/models/tree/master/research/lm_commonsense}} (31 GB) and CC-News~\cite{trinh2018simple} (76 GB) as pretraining corpus. 
For preprocessing, we follow~\citet{deberta} and use the same BPE vocabulary. We use 40 NVIDIA DGX nodes (each with 8$\times$40GB A100 GPU cards) to train our Vega v1 model with a mixed precision training strategy. It takes about 20 days to finish phase-1 (denoising) pretraining with 1 M steps. For phase-2, i.e., contrastive-augmented training, we continuously train Vega v1 for 100K steps. AdamW~\cite{loshchilov2018decoupled} is used as the optimizer for the pretraining stage.
% During fine-tuning, we only apply our self-calibrated fine-tuning strategy to CoLA\footnote{Since it usually suffers from the domain discrepancy between training and test sets, our self-calibrated strategy can effectively alleviate this problem.}. Additionally, we empirically found that the scale-invariant fine-tuning performs better on QNLI, QQP and MNLI tasks. For the other tasks, the vanilla fine-tuning method is used. 

\subsection{Downstream Tasks}
To validate the effectiveness of Vega v1, we use the widely-used GLUE benchmark~\cite{wang2018glue} as the test bed. As one of the most popular NLU benchmarks, GLUE consists of nine challenging NLU tasks, including linguistic acceptability (CoLA,~\newcite{warstadt2019neural}), sentiment analysis (SST-2,~\newcite{socher2013recursive}), paraphrase (MRPC,~\newcite{dolan2005automatically}), textual similarity (STS-B,~\newcite{cer2017semeval}), question paraphrase (QQP), textual entailment (MNLI,~\newcite{williams2018broad}, RTE,~\newcite{giampiccolo2007third}), question-answer entailment (QNLI,~\newcite{rajpurkar2016squad}), and coreference resolution (WNLI,~\newcite{levesque2012winograd}).
% single-sentence tasks (CoLA,~\newcite{warstadt2019neural}, SST-2,~\newcite{socher2013recursive}), similarity and paraphrase tasks (MRPC,~\newcite{dolan2005automatically}, STS-B,~\newcite{cer2017semeval}, QQP\footnote{\url{https://data.quora.com/First-Quora-Dataset-Release-Question-Pairs}}) and inference tasks (MNLI,~\newcite{williams2018broad}, QNLI,~\newcite{rajpurkar2016squad}, RTE,~\newcite{giampiccolo2007third}, WNLI,~\newcite{levesque2012winograd}).
More detailed data statistics for the above tasks can be found in Appendix (Table~\ref{tab:tasks}). 

During fine-tuning, we only apply our self-calibrated fine-tuning strategy to CoLA, as it usually suffers from the domain discrepancy between training and test sets, and our strategy can effectively alleviate this problem. Additionally, transductive fine-tuning is used for WNLI, and scale-invariant fine-tuning is used for QNLI, QQP, and MNLI, while vanilla fine-tuning is for the others. Notably, as suggested by~\citet{liu2019roberta}, for RTE, STS-B, and MRPC tasks, we first fine-tune our Vega v1 model on the MNLI dataset and then continue fine-tuning on their corresponding single-task corpus for better performance.

\subsection{Main Results}
Table~\ref{table_main} reports the final results on the test sets obtained by our Vega v1 and other cutting-edge models on the GLUE benchmark\footnote{We show the detailed ranking results on the GLUE leaderboard in the Appendix. Please refer to Table~\ref{tab:allrank}.}.
As seen, our Vega v1 surpasses the human baselines in terms of average score (91.3 \textit{vs.} 87.1) by a large margin, and achieves new record state-of-the-art performance among four tasks. More encouragingly, Vega v1 is the first to exceed powerful human performance on the two challenging tasks, i.e., SST-2 (human: 97.8\% \textit{vs.} Vega v1: 97.9\%) and WNLI (human: 95.9\% \textit{vs.} Vega v1: 97.9\%). We attribute this success to the efficient pretraining objectives, i.e., denoising and contrastive-augmented objectives. Specifically, the sample-efficient denoising objective used in phase 1 ensures sufficient training of Vega v1. On the other hand, with the help of the proposed contrastive-augmented objective, our Vega v1 can learn better sentence representations, which are beneficial to the downstream language understanding tasks. 

In general, our Vega v1 outperforms the other cutting-edge counterparts and sits atop the GLUE benchmark ranking as of January 1, 2022. These results show the superiority and effectiveness of our model, indicating the significance of more efficient pretraining objectives.

\section{Conclusion}
\label{sec:con}
This paper presents the JD Explore Academy large-scale Vega v1 PLM for the GLUE benchmark. Based on an advanced transformer backbone with disentangled attention, we propose two novel techniques along the pipeline of pretraining and fine-tuning. During the pretraining, we present two efficient pretraining objectives to encourage our Vega v1 model to fully exploit linguistic knowledge and learn better sentence representations. For fine-tuning, a new self-calibrated strategy is proposed to address the domain discrepancy problem though making full use of the model calibration ability. 

We show that these techniques significantly improve the efficiency of model pretraining and the performance achieved on downstream tasks. Specifically, results on the GLUE benchmark show that our Vega v1 model with 1.3 billion parameters achieves state-of-the-art records on 4 out of 9 tasks and ranks first in terms of the macro-average score. Our experience with building Vega v1 demonstrates the necessity of 1) exploring more efficient pretraining objectives, and 2) wisely performing downstream adaptation. 

We also scale the Vega model to a much larger one -- Vega v2~\cite{zhong2022toward} and interestingly find that when scaling up the model size, the current denoising self-supervised objective RTD used in Vega v1 is lightweight but unstable compared to the MLM objective. Therefore, we suggest adopting our training recipes (in Vega v1) for language models smaller than 5 billion, and for larger model scale budgets, the participators could refer to our Vega v2 report~\cite{zhong2022toward}.

\section*{Acknowledgments}
The authors wish to thank the leaderboard maintainer of GLUE for their great efforts in the construction, and their prompt responses. The authors also specially thank Mr. Yukang Zhang (JD Explore Academy), who kindly supports maintaining a stable computing platform. Lastly, the R\&D of the Vega foundation model series could not have been done without the discussions and support of the JDEA-NLP group.

\bibliography{acl2023}
\bibliographystyle{acl_natbib}

\appendix
% Please add the following required packages to your document preamble:
% \usepackage{multirow}
\begin{table*}[]
\caption{\label{tab:ranking} \textbf{Ranking of our submission in terms of average score} in the GLUE leaderboard (\url{https://gluebenchmark.com/leaderboard}) on January 1, 2022.}
\scalebox{0.7}{
\begin{tabular}{clcccccccccccccc}
\toprule
 & \multicolumn{1}{c}{} & CoLA & SST-2 & \multicolumn{2}{c}{MRPC} & \multicolumn{2}{c}{STS-B} & \multicolumn{2}{c}{QQP} & \multicolumn{2}{c}{MNLI} & QNLI & RTE & WNLI &  \\ \cmidrule{3-15}
\multirow{-2}{*}{Rank} & \multirow{-2}{*}{Models} & \textit{Mcc.} & \textit{Acc.} & \textit{F1} & \textit{Acc.} & \textit{Pcor.} & \textit{Scor.} & F1 & Acc. & \textit{m.} & \textit{mm.} & \textit{Acc.} & \textit{Acc.} & \textit{Acc.} & \multirow{-2}{*}{Score} \\ \midrule \midrule
1 & Vega v1 & 73.8 & 97.9 & 94.5 & 92.6 & 93.5 & 93.1 & 76.7 & 91.1 & 92.1 & 91.9 & 96.7 & 92.4 & 97.9 & 91.3 \\
2 & TuringNLR v5 & 72.6 & 97.6 & 93.8 & 91.7 & 93.7 & 93.3 & 76.4 & 91.1 & 92.6 & 92.4 & 97.9 & 94.1 & 95.9 & 91.2 \\
3 & ERNIE & 75.5 & 97.8 & 93.9 & 91.8 & 93.0 & 92.6 & 75.2 & 90.9 & 92.3 & 91.7 & 97.3 & 92.6 & 95.9 & 91.1 \\
4 & StructBERT + CLEVER & 75.3 & 97.7 & 93.9 & 91.9 & 93.5 & 93.1 & 75.6 & 90.8 & 91.7 & 91.5 & 97.4 & 92.5 & 95.2 & 91.0 \\
5 & DeBERT + CLEVER & 74.2 & 97.5 & 93.3 & 91.0 & 93.4 & 93.1 & 76.4 & 90.9 & 92.1 & 91.7 & 96.7 & 93.1 & 96.6 & 91.0 \\
6 & DeBERTa / TuringNLRv4 & 71.5 & 97.5 & 94.0 & 92.0 & 92.9 & 92.6 & 76.2 & 90.8 & 91.9 & 91.6 & 99.2 & 93.2 & 94.5 & 90.8 \\
7 & MacALBERT + DKM & 74.8 & 97.0 & 94.5 & 92.6 & 92.8 & 92.6 & 74.7 & 90.6 & 91.3 & 91.1 & 97.8 & 92.0 & 94.5 & 90.7 \\
8 & ALBERT + DAAF + NAS & 73.5 & 97.2 & 94.0 & 92.0 & 93.0 & 92.4 & 76.1 & 91.0 & 91.6 & 91.3 & 97.5 & 91.7 & 94.5 & 90.6 \\
9 & T5 & 71.6 & 97.5 & 92.8 & 90.4 & 93.1 & 92.8 & 75.1 & 90.6 & 92.2 & 91.9 & 96.9 & 92.8 & 94.5 & 90.3 \\
10 & MT-DNN-SMART & 69.5 & 97.5 & 93.7 & 91.6 & 92.9 & 92.5 & 73.9 & 90.2 & 91.0 & 90.8 & 99.2 & 89.7 & 94.5 & 89.9 \\
11 & NEZHA-Large & 71.7 & 97.3 & 93.3 & 91.0 & 92.4 & 91.9 & 75.2 & 90.7 & 91.5 & 91.3 & 96.2 & 90.3 & 94.5 & 89.9 \\
12 & Funnel-Transformer & 70.5 & 97.5 & 93.4 & 91.2 & 92.6 & 92.3 & 75.4 & 90.7 & 91.4 & 91.1 & 95.8 & 90.0 & 94.5 & 89.7 \\
13 & ELECTRA-Large-large & 71.7 & 97.1 & 93.1 & 90.7 & 92.9 & 92.5 & 75.6 & 90.8 & 91.3 & 90.8 & 95.8 & 89.8 & 91.8 & 89.4 \\
14 & DropAK-ELECTRA-large & 70.4 & 95.8 & 92.6 & 90.1 & 91.2 & 91.1 & 75.1 & 90.5 & 91.1 & 90.9 & 93.8 & 90.1 & 91.8 & 88.7 \\
15 & FreeLB-RoBERTa & 68.0 & 96.8 & 93.1 & 90.8 & 92.3 & 92.1 & 74.8 & 90.3 & 91.1 & 90.7 & 95.6 & 88.7 & 89.0 & 88.4 \\
16 & HIRE-RoBERTa & 68.6 & 97.1 & 93.0 & 90.7 & 92.4 & 92.0 & 74.3 & 90.2 & 90.7 & 90.4 & 95.5 & 87.9 & 89.0 & 88.3 \\
17 & ELECTRA-large-M & 69.3 & 95.8 & 92.2 & 89.6 & 91.2 & 91.1 & 75.1 & 90.5 & 91.1 & 90.9 & 93.8 & 87.9 & 91.8 & 88.3 \\
18 & RoBERTa & 67.8 & 96.7 & 92.3 & 89.8 & 92.2 & 91.9 & 74.3 & 90.2 & 90.8 & 90.2 & 95.4 & 88.2 & 89.0 & 88.1 \\
19 & MT-DNN-ensemble & 68.4 & 96.5 & 92.7 & 90.3 & 91.1 & 90.7 & 73.7 & 89.9 & 87.9 & 87.4 & 96.0 & 86.3 & 89.0 & 87.6 \\
20 & GLUE Human Baselines & 66.4 & 97.8 & 86.3 & 80.8 & 92.7 & 92.6 & 59.5 & 80.4 & 92.0 & 92.8 & 91.2 & 93.6 & 95.9 & 87.1 \\
\bottomrule
\end{tabular}
}
\label{tab:allrank}
\end{table*}
\begin{table*}[t]
\centering \small
\caption{\textbf{Task descriptions and statistics}. All tasks are single sentence or sentence pair classification, except STS-B, which is a regression task. MNLI has three classes; all other classification tasks have two. Test sets shown in bold use labels that have never been made public in any form.
}
\begin{tabular}{lrrlll}
 \toprule
\textbf{Corpus} & \textbf{$|$Train$|$} & \textbf{$|$Test$|$} & \textbf{Task} & \textbf{Metrics} & \textbf{Domain} \\
\midrule
\multicolumn{6}{c}{Single-Sentence Tasks}\\
\midrule
CoLA & 8.5k & \textbf{1k} & acceptability & Matthews corr.& misc. \\ % SB: Changed from 'linguistics literature'. That could be misleading, as few of the sentences are actually in the style of academic writing, and many are found in the wild.
SST-2 & 67k & 1.8k & sentiment & acc. & movie reviews \\
\midrule
\multicolumn{6}{c}{Similarity and Paraphrase Tasks}\\
\midrule
MRPC & 3.7k & 1.7k & paraphrase & acc./F1 & news \\
STS-B & 7k & 1.4k & sentence similarity & Pearson/Spearman corr. & misc. \\
QQP & 364k & \textbf{391k} & paraphrase & acc./F1 & social QA questions \\
\midrule
\multicolumn{6}{c}{Inference Tasks} \\
\midrule
MNLI & 393k & \textbf{20k} & NLI & matched acc./mismatched acc. & misc. \\
QNLI & 108k & 5.7k & QA/NLI & acc. & Wikipedia \\
RTE & 2.5k & 3k & NLI & acc. & misc. \\
WNLI & 634 & \textbf{146} & coreference/NLI & acc. & fiction books \\
\bottomrule
\end{tabular}
\label{tab:tasks}
\end{table*}

\end{document}